\documentclass[10pt,twocolumn,a4paper]{esaAI}

\title{Neural-based Control for CubeSat Docking Maneuvers}

\def\authorEmail{matteo.stoisa@aikospace.com}

\author[1]{Matteo Stoisa\thanks{Corresponding author. E-Mail: \authorEmail}}
\author[1]{Federica Paganelli Azza}
\author[1]{Luca Romanelli}
\author[1]{Mattia Varile}
\affil[1]{AIKO S.r.l., Turin, Italy}

\usepackage{xcolor}
\usepackage{multirow}
\usepackage{soul}

\begin{document}

\makeCustomtitle

\begin{abstract}

Autonomous Rendezvous and Docking (RVD) have been extensively studied in recent years, addressing the stringent requirements of spacecraft dynamics variations and the limitations of GNC systems. This paper presents an innovative approach employing Artificial Neural Networks (ANN) trained through Reinforcement Learning (RL) for autonomous spacecraft guidance and control during the final phase of the rendezvous maneuver. The proposed strategy is easily implementable onboard and offers fast adaptability and robustness to disturbances by learning control policies from experience rather than relying on predefined models. Extensive Monte Carlo simulations within a relevant environment are conducted in 6DoF settings to validate our approach, along with hardware tests that demonstrate deployment feasibility. Our findings highlight the efficacy of RL in assuring the adaptability and efficiency of spacecraft RVD, offering insights into future mission expectations.
\end{abstract}

\section{Introduction}
Rendezvous and Docking (RVD) operations are critical for a wide range of space missions, enhancing capabilities such as inspection, observation, active debris removal, and space tug operations \cite{arney2023}. These operations are essential for maintaining and enhancing space assets' operational lifespan and functionality.

The adoption of small satellites for rendezvous missions presents unique challenges due to their reduced size and the limited available technologies. While recent advancements have brought small satellite technology to the required level of maturity, there is still a lack of adaptive guidance and control solutions \cite{wood2023}. RVD missions require precise maneuvers for the chaser spacecraft to dock with the target, with the final meters being especially critical due to the reduced margin for error \cite{fehse2003}.

This paper focuses on an innovative method for RVD of small satellites, leveraging Artificial Neural Networks (ANN). This approach is designed to implement the control strategies for the chaser spacecraft, addressing the challenges posed by the small size and operational constraints of CubeSats. By exploiting ANN, the guidance and control systems can adapt to dynamic conditions and uncertainties, improving the overall performance of the docking maneuvers. Moreover, the computational cost associated to such Neural Networks (NN) is much lighter compared to traditional model-based control methods such as Model Predictive Control (MPC) \cite{mpc}. We believe this is the main motivation for advancing the development of this type of technology applied to CubeSats, where the on-board computational load is a major constraint having limited power availability.

\subsection{Deep Reinforcement Learning for control}
\label{sec:goal}

Recently, the use of ANN has matched state-of-the-art performance in accomplishing control tasks \cite{Kaufmann2023} by successfully overcoming the duality between data-driven training through simulations and deployment in the real world. The exploit of a single software that maps real-time navigation data into control signals could offer computational and performance improvements, although it challenges the historical approach that distinguishes guidance generation and controller \cite{Song_2023}. This technology is also identified in the literature as G\&CNET \cite{izzo}. While its validity has been demonstrated through optimal real-world drone control \cite{Kaufmann2023}, various studies are evaluating its applicability to space scenarios \cite{izzo_2024} such as pinpoint landing \cite{sanchez-landing} \cite{furfaro-landing}, low-thrust rendezvous \cite{izzo} and spacecraft orbit transfer \cite{izzo-trajectories}.

The supervised approach of Imitation Learning \cite{IL} bases training procedures on pre-computed databases represented by optimal expert demonstrations to obtain policies that mimic the reference behavior. On the other hand, the Deep Reinforcement Learning (DRL) framework \cite{Sutton1998} bases the learning process on experience exploiting a goal-oriented approach. \textit{Agents} can learn to solve disparate tasks by directly interacting with the dynamic environment, only the objectives and the variables subject to optimization have to be specified, while the training process itself discovers information on how to improve the behavior towards optimality.

In this context, we aim to demonstrate the feasibility of training through DRL such neural controller applied to the RVD use case. We perform validation through Monte Carlo simulations in a high-fidelity environment and conduct preliminary tests on space-relevant platforms. To build an ANN able to accomplish the task, our strategy is to gradually inject noise via stochastic processes within the training iterations so that the final \textit{policy} is able to handle non-nominal scenarios consistently.

\section{Problem formalization}
\label{sec:sec2}

The reference scenario for this study involves the final approach phase of a docking mission between two identical 6U CubeSats with a mass of 12$kg$. Specifically, the control phase begins at a final hold point located 15$m$ from the target spacecraft, with the docking port aligned with the chaser. Throughout the maneuver, the chaser's attitude is maintained to point toward the target's Center of Mass (CoM). The docking interfaces are assumed to coincide with the CoMs.

The chaser is equipped with a set of 24 cold-gas thrusters, each of them characterized by a maximum thrust $T_{max}$ of $10mN$ and a specific impulse $I_{sp}$ of $60s$. It operates in a near-circular Low Earth Orbit (LEO) and is tasked with docking to a stationary target spacecraft with a known docking port position. The initial conditions of the docking maneuver are defined according to a V-bar approach \cite{fehse2003}.

The goal of the approach introduced in \cref{sec:goal} is to generate directly onboard a translational 3DoF control profile for the chaser. In particular, the trajectory must lie inside a safety cone with a half-angle of 10° and the soft docking performance described in \cref{tab:soft_docking} must be reached.

\begin{table}[h]\renewcommand{\arraystretch}{1.2}
    \begin{center}
        \begin{tabular}{c c} 
        \hline\hline
         \textbf{Quantity} & \textbf{Required performance} \\
        \hline\hline
        Lateral alignment $[m]$ & $< 0.005$ \\
        Approach velocity $[m/s]$ & $< 0.02$ \\
        Normal velocity $[m/s]$ & $< 0.01$ \\
        \hline\hline
        \end{tabular}
    \caption{Docking requirements.}
    \label{tab:soft_docking}
    \end{center}
\end{table}

\section{Methodology}
\label{sec:sec3}

\begin{figure*}[t]
    \centering
    \includegraphics[width=.75\textwidth]{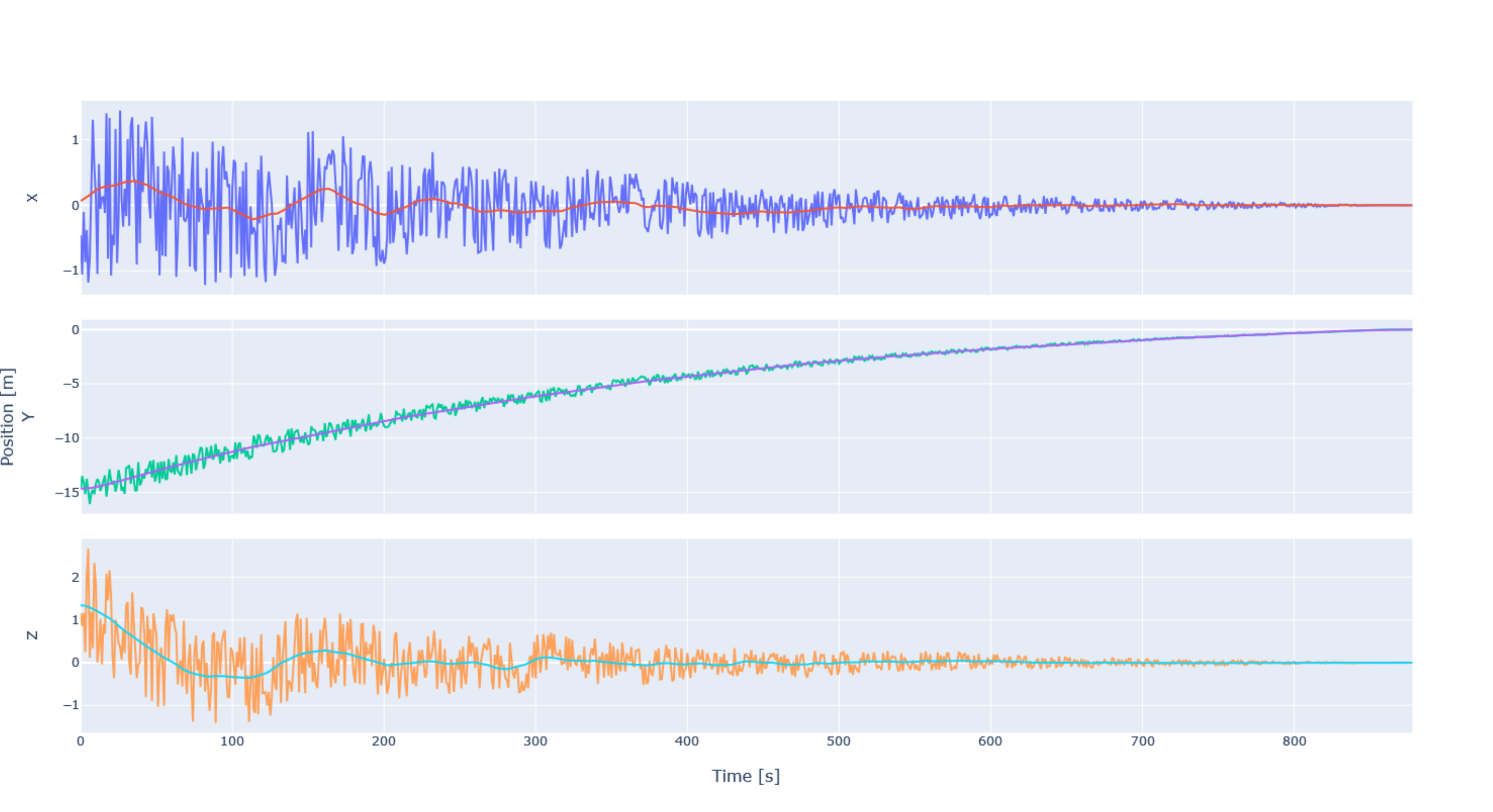}
	\caption{Observations (back) and real values (front)  of position coordinates during a docking maneuver using $\alpha=0.1$.}
	\label{fig:observation}
\end{figure*}

In this section, we describe the simplified chaser's translational dynamics exploited during the training phases, along with spacecraft and perturbation modeling, and the RL setup building our approach. The validation will employ a more complex model of the translational dynamics.

\subsection{Spacecraft model}

A simplified set of equations exists in the case of very low relative chaser-target distance to orbit radius ratio, while also assuming a circular orbit \cite{sidi1997}. The equations of motion are reported in \cref{eq:hill}, considering the Local Vertical Local Horizontal (LVLH) reference frame centered in the target CoM.:
\begin{equation} \label{eq:hill}
    \begin{split}
       \ddot{x} =& \: 2\Omega\dot{y}+3\sigma^2x+\frac{F_x}{m} \\
       \ddot{y} =& \: -2\Omega\dot{x}+\frac{F_y}{m} \\
       \ddot{z} =& \: -\Omega^2z+\frac{F_z}{m}
    \end{split}
\end{equation}
where $x, y, z$ are the relative position coordinates, $F_x, F_y, F_z$ are the chaser's thrust forces, $m$ is the mass of the chaser, and $\Omega$ is the orbital rate of the target body in its circular orbit.

\subsection{Reinforcement Learning setup}

To perform the 3DoF control exploiting an ANN trained through the RL framework, we formalize the problem as a Markov Decision Process (MDP) \cite{Sutton1998}. The observation space is composed of distances and velocities relative to the target CoM in $x, y, z$ frame, and the mass. These values are normalized and clipped. Three continuous values within $[-1,1]$ make up the action space, then mapped within $[-T_{max}, T_{max}]$, each representing the amount of thrust throttle requested in a direction. The sum of these 3 components is the force vector imposed in the CoM. The control step is set to 1$s$, the task can last a maximum of 1000$s$.

We use Proximal Policy Optimization (PPO) \cite{schulman2017proximal} as a Reinforcement Learning algorithm to train a multilayer perceptron (MLP) that is in charge of performing the 3DoF control. The MLP, generically called ANN, contains two internal layers of 256 hidden units and a terminal \textit{tanh} activation function. The main algorithm hyperparameters for each training iteration are specified in \cref{tab:algo}.
\begin{table}[h]\renewcommand{\arraystretch}{1.2}
    \begin{center}
        \begin{tabular}{ c c c c } 
        \hline\hline
        \textbf{$\gamma$} & \textbf{lr} & \textbf{batch size} & \textbf{training steps} \\
        \hline\hline
        0.999 & $5 \cdot 10^{-5}$ & 64 & $25 \cdot 10^6$ \\
        \hline\hline
        \end{tabular}
        \caption{Main PPO hyperparameters.}
        \label{tab:algo}
    \end{center}
\end{table}

The scenario described poses numerous difficulties that must be addressed during a learning process that evolves from scratch, not based on expert experiences. Primarily, the precision requirements on the success condition complicate the exploration necessary to achieve the goal, which is surrounded by failure states of collision with the safety cone or the target. In addition, trade-off has to be acceptably managed between the minimization of fuel consumption and maneuver duration, avoiding too "lazy" or "aggressive" behaviors. Also, other than being generalizable to initial conditions, we want the solution to be robust to noise on both input and output signals, formalized in the next section. The exploitation of different reward functions and Curriculum Learning \cite{curriculum_rl} strategies demonstrated crucial to overcome these challenges, allowing the learning process to first discover the task success and then to optimize its achievement.

\subsection{Simulations} \label{subsec:subsec3.3}

\begin{figure*}[t]
    \centering
    \includegraphics[width=.75\textwidth]{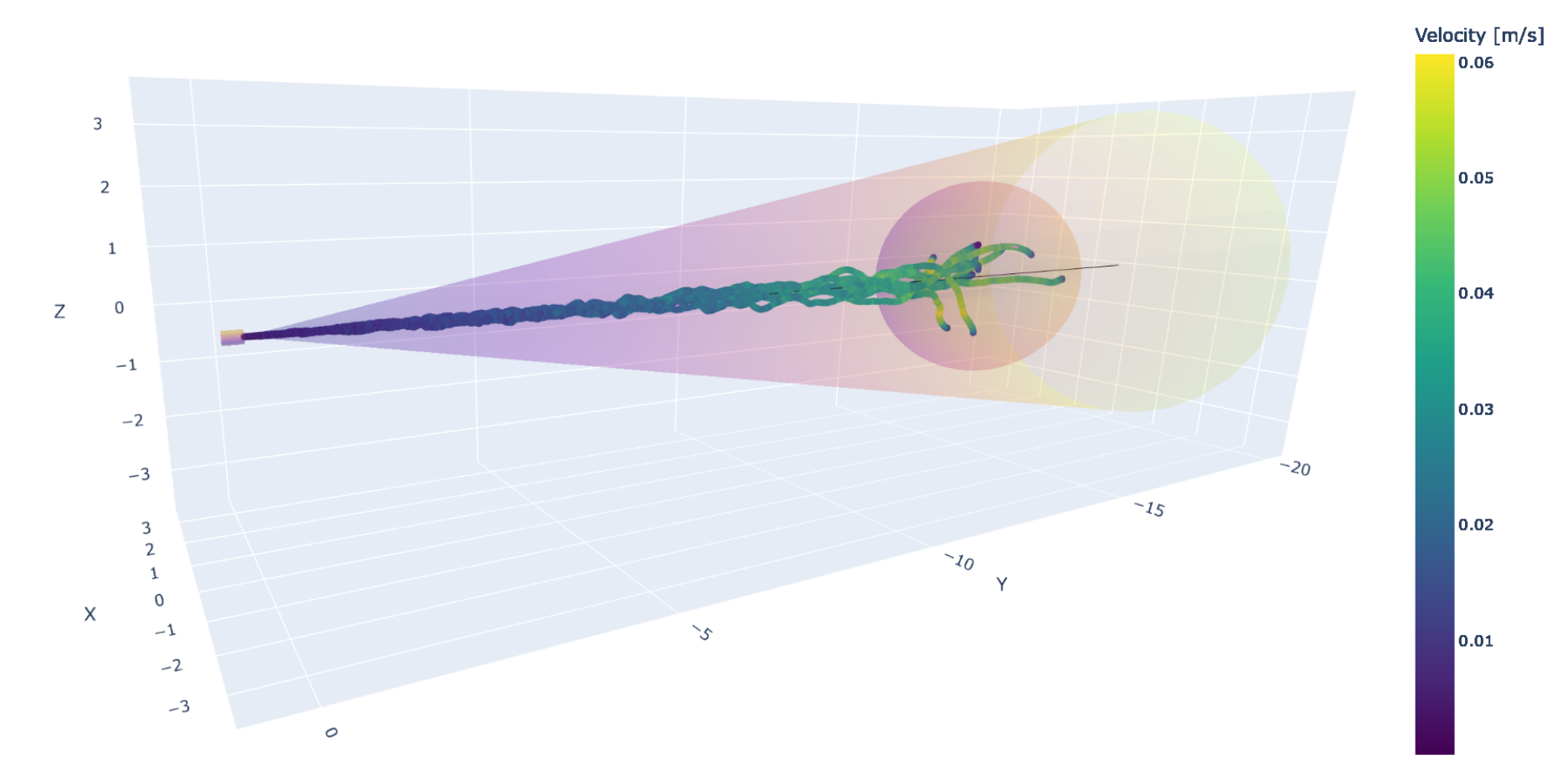}
	\caption{Visualization of 10 maneuver trajectories using $\alpha=0.05$. Other than the safety cone and its axis, the sphere where initial conditions are drawn is displayed.}
	\label{fig:trajectories}
\end{figure*}

The physical conditions with which training and validation simulations start are sampled from uniform distributions in order to assure adaptability to stochastic configurations of a possible real mission. \cref{tab:initial} shows distributions of parameters summed to the nominal values to initialize each mission; position and velocity deltas are applied in uniformly random directions.
\begin{table}[h]\renewcommand{\arraystretch}{1.2}
    \begin{center}
        \begin{tabular}{ c c } 
        \hline\hline
        \textbf{Quantity} & \textbf{Displacement distribution} \\
        \hline\hline
        Position $[m]$ & $\mathcal{U}(-2, 2)$ \\
        Velocity $[m/s]$ & $\mathcal{U}(-0.02, 0.02)$ \\
        Mass $[kg]$ & $\mathcal{U}(-1.2, 1.2)$ \\
        \hline\hline
        \end{tabular}
        \caption{Distributions of initial condition displacements.}
        \label{tab:initial}
    \end{center}
\end{table}
To make the controller's behavior robust to conditions that would not occur in nominal simulations, we inject stochastic uncertainties and anomalies into both training and validation procedures, respectively, in the navigation data and the actuation subsystem.

To implement thruster non-nominalities, starting from the nominal force vector selected by the ANN, at each step we instead apply a perturbed thrust, introducing noise in the magnitude and direction. Let the requested vector be the nominal thrust $\boldsymbol{T}$ of magnitude $T \in [0, T_{max}]$ and direction identified by spherical angles $\phi \in [0, 2 \pi]$ and $\theta \in [0, \pi]$. The perturbed action $\boldsymbol{\hat{T}}$ is composed by perturbed magnitude $\hat{T}$ and perturbed direction $\hat{\theta}$, defined as:
\begin{equation} \label{eq:magnitude}
    \begin{split}
        \hat{T} =& \: clip(0, \ T_{max}, \ T + \delta_{T} T_{max}) \\
        with \:\:\: \delta_{T} \sim& \mathcal{U}(0, \chi T_{max})
    \end{split}
\end{equation}
\begin{equation} \label{eq:direction}
    \begin{split}
        \hat{\theta} =& \: rod(\theta + \delta_{\theta}, \theta, \zeta) \\
        with \:\:\: \delta_{\theta} \sim& \mathcal{U}(0, \iota),
        \zeta \sim \: \mathcal{U}({-\pi, \pi})
    \end{split}
\end{equation}
We use $\chi=0.0525$ and $\iota=0.1^{\circ}$ to model the actuation uncertainties according to off-the-shelf electric thrusters for CubeSats. The $clip$ function prevents the perturbed magnitude from resulting in a negative or greater than the maximum deliverable thrust. $rod(v_{1}, v_{2},\zeta)$ is the Rodrigues' rotation {\cite{rodrigues}} for vector $v_1$ around $v_2$ of an angle $\zeta$. In other words, the nominal direction $\theta$ is deflected by an angle $\delta_{\theta}$ in a random direction.

To model errors in data coming from the navigation pipeline, we inject stochastic noise in the observation values of positions and velocities. Given at each step $t$ the exact measurements $o_t$ for each axis, we provide as input to the ANN perturbed observations $\hat{o}_t$ computed independently as:
\begin{equation} \label{eq:observation}
    \begin{split}
        \hat{o}_{t} =& \: o_{t} + \delta_{o_t} \\
        with \:\:\: \delta_{o_t} \sim& \: \mathcal{U}(-\frac{d_{t}}{d_{max}} \alpha, \frac{d_{t}}{d_{max}} \alpha) \\
    \end{split}
\end{equation}
where $d_t$ is the Euclidean distance from the chaser and target CoMs, and $d_{max}=20m$. According to this model, the observation noise decreases linearly with the distance from the target, and it is symmetric and centered around the real value. While characterizing the errors of a pose estimation algorithm is beyond the scope of this paper, this noise implementation represents the filtered outputs of a relative navigation pipeline suitable for CubeSats \cite{BECHINI202420}. The hyperparameter $\alpha$ represents the navigation pipeline's reliability, $\alpha=0$ corresponds to optimal observations while $\alpha=0.1$ means that at maximal distance $d_{max}$ the noise displacement is up to 10\% of the real measurement. \cref{fig:observation} shows an example of position components perturbed and nominal measurements.

The observation of the mass value $m_t$ is perturbed at each step as follows:
\begin{equation} \label{eq:mass}
    \begin{split}
        \hat{m}_{t} =& \: m_{t} + \delta_{m} \\
        with \:\:\: \delta_{m} \sim& \: \mathcal{U}(-m_i \alpha, m_i \alpha) \\
    \end{split}
\end{equation}
where $m_i$ is the initial mass.

\section{Results}

The work aims to verify and quantify the NN-controller's robustness and performance within a validation environment that offers high-fidelity dynamics modeling. Details regarding training metrics and the comparison between training and validation are omitted for brevity, treating only the behavior achieved.

\begin{table}[h]\renewcommand{\arraystretch}{1.2}
    \begin{center}
        \begin{tabular}{ c c  c  c } 
        \hline\hline
         \textbf{$\alpha$} & \textbf{Success $[\%]$} & \textbf{$\overline{\Delta m}$ $[g]$} & \textbf{$\overline{\Delta t}$ $[s]$} \\
        \hline\hline
        0 & 100.0 & $2.968 \pm 0.602$ & $612.2 \pm 14.9$ \\
        0.05 & 100.0 & $63.545 \pm 1.692$ & $909.6 \pm 33.4$ \\
        0.1 & 99.9 & $77.489 \pm 1.402$ & $903.1 \pm 17.2$ \\
        \hline\hline
        \end{tabular}
    \caption{Validation results.}
    \label{tab:results}
    \end{center}
\end{table}

\begin{figure*}[t]
    \centering
    \includegraphics[width=.75\textwidth]{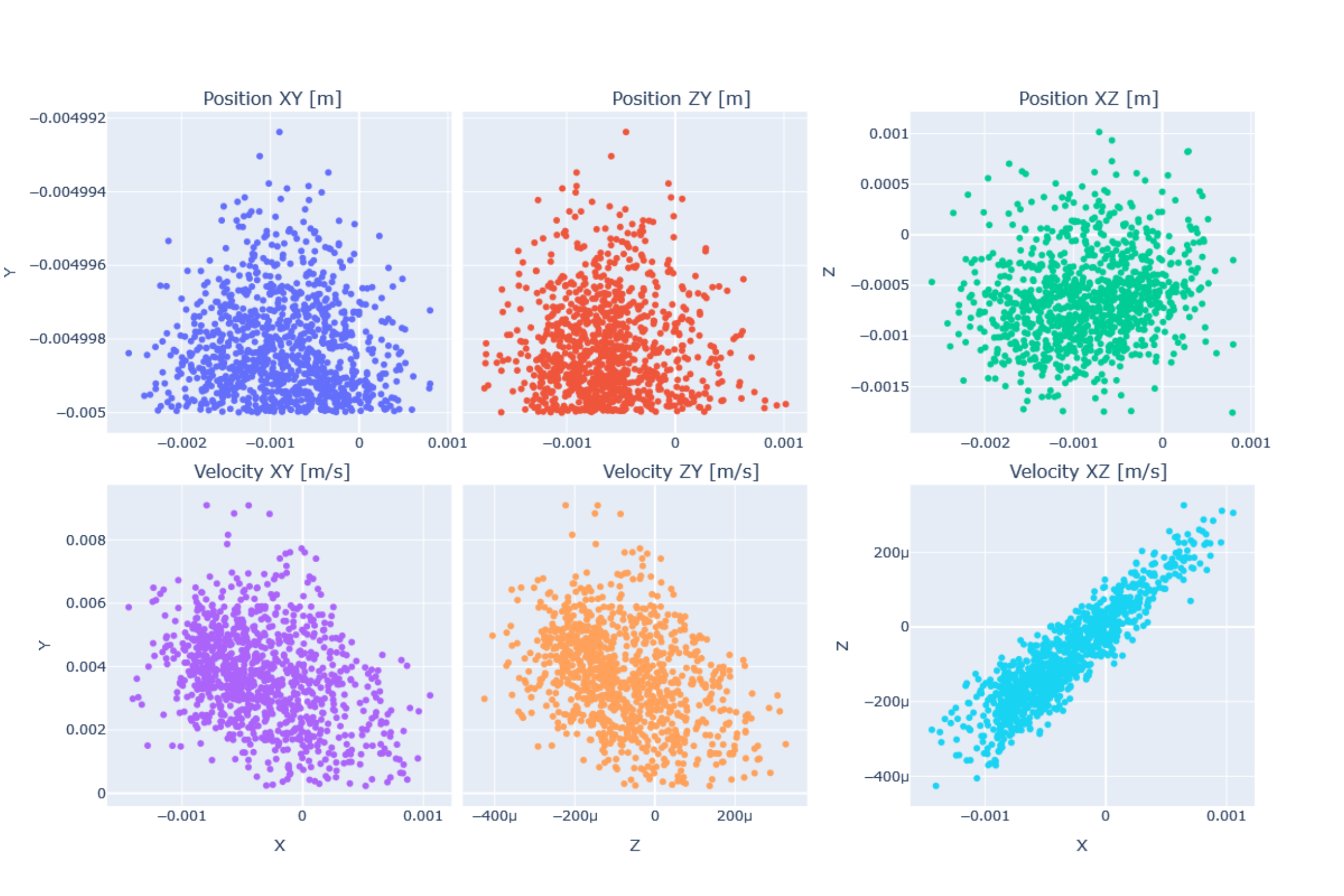}
	\caption{Terminal position and velocity values using $\alpha=0.05$. Regarding distances reached along the Y-axis, the -0.005$m$ margin represents the minimal distance from the target to consider the docking satisfied. }
	\label{fig:terminations}
\end{figure*}

We validate our Agent through Monte Carlo simulations within \textit{42} \cite{Stoneking10}, an open-source spacecraft simulator by NASA that implements a nonlinear translational model. An ad-hoc software interface allows the closed-loop control between \textit{42} and our Agent inference engine. In addition to the relative orbit control approach discussed earlier, we utilize the Proportional-Derivative (PD) attitude controller provided by \textit{42} to achieve 6DoF control in a decoupled mode. While this controller's performance may not be optimal, our tests aim to address challenges related to error management and delays introduced by the attitude control system. As validation of the proposed approach, we analyze 1000 Monte Carlo simulations performed in \textit{42} with incremental $\alpha$. Statistics of these maneuvers and visualization of terminal states are shown in \cref{tab:results} and \cref{fig:terminations}, in the latter it can be seen that tolerances given in \cref{tab:soft_docking} are largely satisfied.

\subsection{Hardware tests}

To explore the feasibility of deploying this technology onboard, we conduct inference tests using two different embedded GPU NVIDIA$^{\circledR}$ modules: Jetson Nano and Jetson Xavier. Radiation-tolerant devices based on these accelerators have already been tested in orbit. We employ both CPU and GPU runtime, respectively using \textit{ONNX} and \textit{TensorRT}, with \textit{16-floating-point} precision while Monte Carlo simulations previously described are performed with \textit{32-floating-point} precision. Mean inference times, reported as $\tau$ in \cref{tab:hw}, demonstrate that CPU runtime is much faster; this is understandable considering the small size of the NN and the single batch size. The inference duration is one order of magnitude shorter than the control period, supporting the feasibility of the application in a deployment environment without requiring ad-hoc hardware accelerators.

\begin{table}[h]\renewcommand{\arraystretch}{1.2}
    \begin{center}
        \begin{tabular}{ c  c  c } 
        \hline\hline
         \textbf{Device} & \textbf{Runtime} & \textbf{$\overline{\tau}$ $[ms]$} \\
        \hline\hline
        \multirow{2}{*}{Jetson Nano} & ONNX (CPU) & $0.093 \pm 0.007$ \\
        & TensorRT (GPU) & $0.516 \pm 0.085$ \\
        \multirow{2}{*}{Jetson Xavier} & ONNX (CPU) & $0.145 \pm 0.028$ \\
        & TensorRT (GPU) & $0.747 \pm 0.028$ \\
        \hline\hline
        \end{tabular}
    \caption{Hardware tests statistics.}
    \label{tab:hw}
    \end{center}
\end{table}

\section{Discussion}

Given results shown in \cref{tab:results}, the objective of adaptability and robustness in transitioning from the training environment to \textit{42} are demonstrated. Stochastic initial conditions and disturbances on actuation and navigation are handled robustly, as shown in \cref{fig:terminations}, where the dispersion of measurements is relatively narrow considering the strict requirements. As might be expected, performance with $\alpha=0$ is considerably better, while the trend of average resource consumption when increasing observation noise is consistent.

Based on the simulation results and preliminary platform tests, we demonstrated that this technology reaches a satisfactory level of generalizability with limited computational cost. The next step is to assess the applicability and performance of this approach in scenarios involving uncooperative targets with tumbling motions. In these scenarios, we expect the designed policy to showcase its advantages over classical methodologies, particularly in its ability to generalize effectively even in the presence of unmodeled dynamics.

\clearpage

\printbibliography
\addcontentsline{toc}{section}{References}

\end{document}